\PassOptionsToPackage{table}{xcolor}
\documentclass[]{fairmeta}
\microtypesetup{expansion=false}
\usepackage{amsmath}
\usepackage{amssymb}
\usepackage{mathtools}
\usepackage{colortbl}
\usepackage{tabularx}
\usepackage{enumitem}

\providecommand{\Description}[1]{}
\newcommand{\worldscope}{WorldScope}
\newcommand{\worldset}{WorldScope-1.2M}
\newcommand{\worldbench}{WorldScope-Bench}
\newcommand{\worldflow}{WorldFlow}

\definecolor{wfheader}{RGB}{255,255,255}
\definecolor{wfapi}{RGB}{232,243,252}
\definecolor{wfopen}{RGB}{235,247,234}
\definecolor{wfhuman}{RGB}{243,237,250}
\definecolor{wfblue}{RGB}{232,243,252}
\definecolor{wfgreen}{RGB}{255,239,220}
\definecolor{wfgray}{RGB}{255,255,255}
\definecolor{wfbest}{RGB}{180,0,0}
\definecolor{wfred}{RGB}{250,229,225}
\definecolor{wfgreenhead}{RGB}{255,231,202}
\definecolor{wfink}{RGB}{62,58,55}
\newcommand{\wfmainhead}[1]{{\fontsize{8.4}{10}\selectfont\shortstack{#1}}}
\newcommand{\wfhead}[1]{{\fontsize{9}{10.5}\selectfont\shortstack{#1}}}
\newcommand{\best}[1]{\textcolor{wfbest}{\textbf{#1}}}
\newcolumntype{Y}{>{\centering\arraybackslash}X}
\newcolumntype{G}{>{\columncolor{wfgreen}\centering\arraybackslash}X}
\newcolumntype{L}[1]{>{\raggedright\arraybackslash}p{#1}}
\newcolumntype{Z}{>{\raggedright\arraybackslash}X}

\title{Seeing Parts, Reasoning about Worlds: Visual Inference under Partial Observation}
\author{Wei Wang}
\author[*]{Wenqiao Zhang}
\author{Yutong Lin}
\author{Jun Xiao}
\author{Yueting Zhuang}
\affiliation{College of Computer Science and Technology, Zhejiang University}
\contribution[*]{Corresponding author.}

\abstract{
World modeling under partial observation requires reasoning about the complete
worlds that remain compatible with limited visual evidence.  Occluded objects
and unseen regions can leave several world states possible; additional views
can exclude alternatives and strengthen the conclusions supported by the observations.
We introduce \worldscope{} to study this process through possible-world
semantics, evidence-grounded data, and learned visual representations.
\worldset{} provides 1.2 million English question--answer pairs spanning eight
world properties, ten task interfaces, and three observation protocols.
Its answers encode confirmed facts, supported bounds, and unresolved possibilities.
Complementary supervision comprises 4,800 certified counterworld groups with
equivalent base observations and different hidden object configurations and
query answers.  These groups provide physical witnesses of ambiguity and
training-only labels for world compatibility and view-induced exclusions.
We propose \worldflow{}, which composes cross-view entity evidence and
support-surface coverage into an image-subset evidence lattice.
Counterworld compatibility and transition objectives train subset representations
to reflect how new observations constrain possible worlds.  A shared answer
generator uses these representations to predict the strongest supported conclusion.
\worldbench{} evaluates claim judgments and evidence-dependent conclusions as
views are selected, combined, removed, or ordered.  On its 5,000-question test
set, \worldflow{} reaches 64.34\% exact accuracy, improving over the same
backbone trained on QA alone by 24.88 percentage points.  It retains 50.43\%
accuracy on the 3,000 questions from structure-disjoint scenes.
}

\metadata[Keywords]{world modeling, possible-world reasoning, partial observation, visual inference, visual identifiability}
\hypersetup{pdftitle={Seeing Parts, Reasoning about Worlds: Visual Inference under Partial Observation},pdfauthor={Wei Wang, Wenqiao Zhang, Yutong Lin, Jun Xiao, Yueting Zhuang},pdfkeywords={world modeling, possible-world reasoning, partial observation, visual inference, visual identifiability}}
\begin{document}
\maketitle

\begin{figure}[tp]

  \centering
  \includegraphics[width=\textwidth]{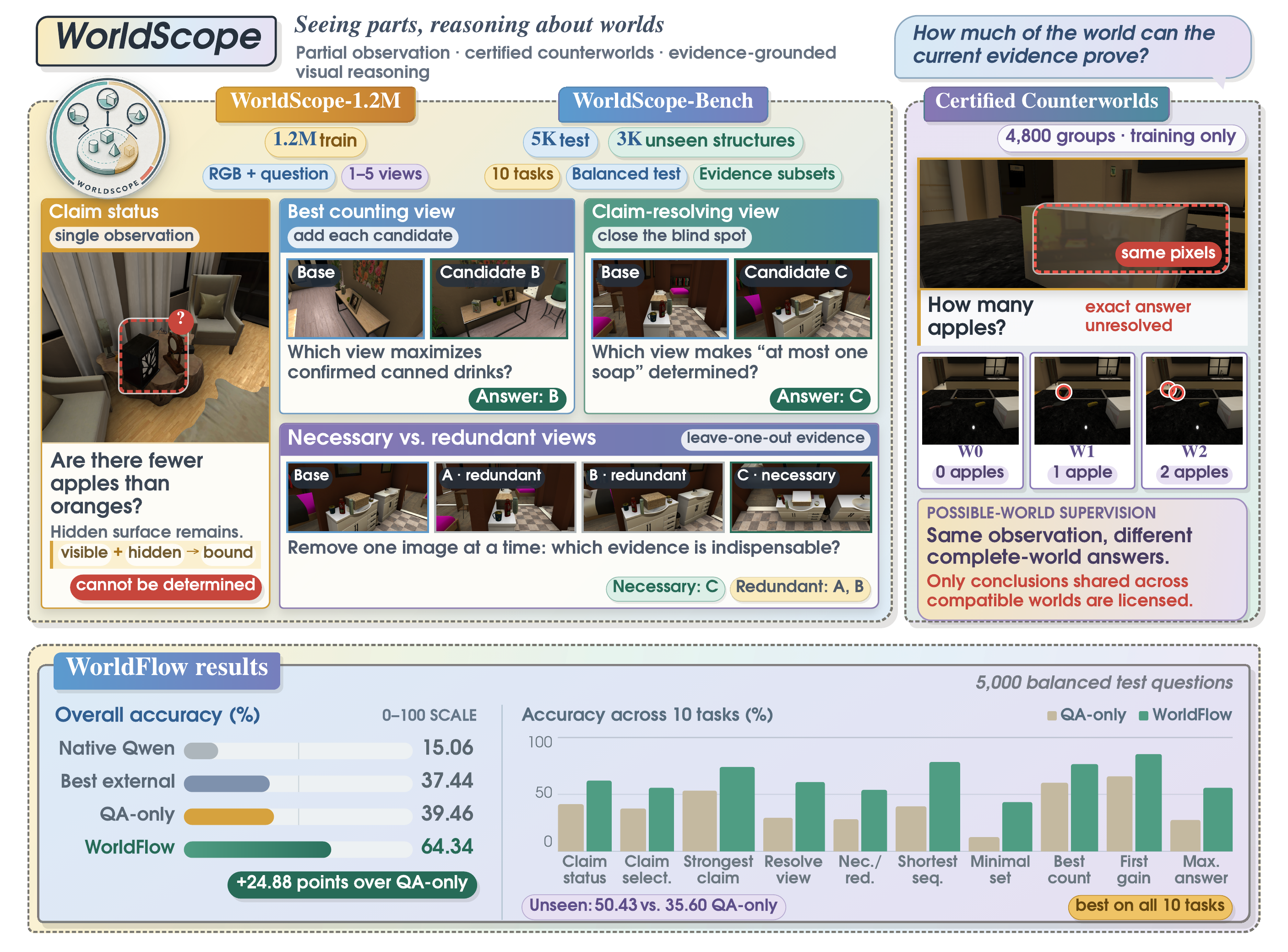}
  \caption{Overview of \worldscope{}.  Partial RGB observations may support a
  bound or leave a claim unresolved even when visible objects are recognized.
  \worldset{} and \worldbench{} cover eight world properties and ten task
  interfaces, while certified counterworlds expose equivalent observations
  with different hidden answers.  The lower panels compare overall and
  task-wise accuracy: \worldflow{} reaches 64.34\%, a 24.88-point gain over
  the same-backbone QA-only baseline.}
  \Description{An overview of WorldScope with claim-status, counting-view,
  resolving-view, and view-necessity examples beside a certified counterworld
  group.  The lower section compares overall and task-wise model accuracy.}
  \label{fig:teaser}

\end{figure}

\section{Introduction}
\label{sec:intro}

Visual observations reveal only part of a world; occluded regions and content outside the field of view
remain unknown.  Consider a tabletop image containing two visible objects of
the same category while another object hides part of the surface.  The evidence
supports ``at least two,'' but it does not justify ``exactly two,'' because the
hidden region may contain another instance.  A partially observed scene can
therefore admit multiple complete configurations consistent with the visible
evidence.  We call each such configuration a \emph{possible world}.  Different
possible worlds may yield different exact answers while sharing weaker
conclusions.  If the total count could be two, three, or four, for example, the
model can still conclude that it is at least two.  Visual reasoning under
partial observation thus requires extracting the strongest conclusion shared
by the possible worlds compatible with the current evidence, including exact
values, bounds, existence claims, and ``cannot be determined'' when the
requested conclusion remains unresolved.

Additional observations further restrict the compatible possible worlds.  A
complementary image can expose a previously hidden region, exclude some
configurations, and tighten a weak conclusion into an exact answer.  A repeated
or irrelevant image may increase the visual input without excluding any new
world.  Single-image judgment and multi-image reasoning consequently share the
same central problem: determine which possible worlds remain compatible with
the current observations and summarize the strongest conclusion they jointly
support.  Figure~\ref{fig:teaser} summarizes how \worldscope{} connects this
evidence semantics to data construction, evaluation, and model design.

Modern multimodal large language models (MLLMs) can process interleaved images
and text and increasingly support cross-image comparison and spatial reasoning
~\cite{li2024llavanextinterleave,jiang2024mantis,bai2025qwen3vl}.  Assessing how
strongly these observations support a conclusion about the whole world remains
a separate requirement.  Three limitations motivate our study.

\textbf{First, existing data overlook hidden possibilities.}
Visual question answering datasets typically ask about content directly visible
in an image.  Synthetic-scene datasets often use a property of the complete
scene as the answer~\cite{antol2015vqa,hudson2019gqa}.
Both generally provide one target answer without labeling which hidden
configurations remain possible beyond the image or distinguishing exact
answers, bounds, and unresolved conclusions.  Supervision for reasoning about
unobserved content therefore remains limited.

\textbf{Second, existing benchmarks do not test evidence boundaries.}
Visual entailment and multi-image benchmarks mainly compare predictions against
preset answers under a fixed input, evaluating evidence judgment, cross-image
comparison, or scene integration~\cite{xie2019visualentailment,suhr2019nlvr2,mo2025mvscanqa}.
They do not systematically test whether a model keeps its conclusions within
what the current evidence permits, or whether answers change correctly as
observations are added, removed, and recombined.  Evaluation under partial
observation should test whether a model derives the strongest supported
conclusion while retaining uncertainty about hidden content.

\textbf{Third, related research lacks a unified model of possible-world
constraints.}
Multi-image models fuse continuous features, cross-view methods establish
object correspondence, and occlusion or coverage methods characterize unseen
regions.  Reasoning under partial observation requires integrating these
components to represent repeated objects across images, regions that may still
contain hidden objects, and the contraction of possible worlds as images are
combined.

We address these gaps with \worldscope, comprising two complementary sources of
training supervision, the structure-disjoint \worldbench, and the \worldflow{}
model.  Together, they support the study of which conclusions a model can
reliably draw from partial observations.

On the data side, \worldset{} provides 1.2 million English question--answer
pairs over 20 HSSD structures and 240 rendered scene instances, spanning eight
world properties, ten task interfaces, and three observation
protocols~\cite{khanna2024hssd}.  Its questions and answers are derived jointly
from the complete scene, visible objects, and uncovered regions, so they capture
both confirmed facts and the bounds or uncertainty caused by unseen space.  The
805,955 distinct reasoning instances use stable identities to prevent duplicate
cross-view counting and dense, object-size-aware surface probes to determine
whether an unseen region can contain the queried object.  A complementary
corpus contains 4,800 certified counterworld groups and 14,400 complete worlds.
Worlds in a group share the same base observation but differ in hidden object
configuration, directly labeling which worlds remain compatible and which are
excluded by a new image.  These worlds provide only visual compatibility
supervision; all natural-language questions remain drawn from \worldset{}.

On the evaluation side, \worldbench{} tests whether a model can derive the
strongest conclusion justified by a partial observation.  The model must use
visible facts while preserving possibilities left by unseen regions, avoid
overclaiming when evidence is insufficient, and update its answer when images
are added, removed, or combined.  Ten task interfaces cover single-image
judgment, multi-image composition, and active disambiguation.  The test set
contains 5,000 balanced questions, including 3,000 from three HSSD structures
entirely unseen during training.

On the model side, \worldflow{} follows the formation of evidence from
individual images to image combinations.  The visual features of each image are
first compressed into 64 per-image evidence vectors.  For every non-empty image
subset, a shared entity fusion module uses 16 slots to merge repeated physical
objects across images, while a coverage branch estimates a $4\times4$ map of
the support-surface regions observed by the subset.  A subset encoder then compresses the entity slots,
coverage estimate, and member-image evidence into four subset evidence vectors.
The language model receives these vectors together with the original images and
question to generate one answer.  Entity and coverage supervision distinguish
confirmed content from regions that may still hide an object, while certified
counterworld supervision teaches the subset vectors how the range of possible
worlds contracts as observations are added.

Our main contributions are:
\begin{itemize}[leftmargin=1.5em]
  \item We introduce the \worldscope{} data suite, in which \worldset{} provides
  1.2 million natural-language question--answer pairs and 4,800 certified
  counterworld groups provide training-only world-compatibility supervision
  without creating a second language-question distribution.
  \item We build \worldbench, which tests whether models derive the strongest
  conclusion supported only by a partial observation and update that conclusion
  correctly as visual evidence is acquired, removed, and composed.
  \item We propose \worldflow, which forms a hierarchical evidence
  representation through per-image compression, cross-view entity fusion,
  coverage prediction, and image-subset encoding, and uses certified
  counterworlds to learn how observations contract the range of possible worlds.
\end{itemize}

\section{Related Work}
\label{sec:related}

\subsection{Multi-Image Reasoning}

Early general-purpose vision--language models connected frozen or pretrained
visual encoders to large language models through resampling, query, or
projection modules~\cite{alayrac2022flamingo,li2023blip2,liu2023visual,dai2023instructblip}.
Qwen-VL, Qwen2-VL, and Qwen3-VL extended this interface to localization,
dynamic visual resolution, videos, and long interleaved multimodal
contexts~\cite{bai2023qwenvl,wang2024qwen2vl,bai2025qwen3vl}.
LLaVA-NeXT-Interleave and MANTIS showed that dedicated interleaved instruction
data can substantially improve comparison, reference, and reasoning across
multiple images~\cite{li2024llavanextinterleave,jiang2024mantis}.  These systems
establish a general multi-image interface and evaluate whether a model can
combine content across images.  Their supervision is usually attached to
the final answer after all images have been encoded into
one context.  A correct answer therefore need not reveal which image supplied
new evidence, whether a smaller subset was already sufficient, or whether the
answer claims more than the visible evidence supports.

Spatially focused MLLMs add metric, region, depth, or three-dimensional
supervision.  SpatialVLM automatically generates large-scale spatial QA from
detected objects and estimated depth, whereas SpatialRGPT grounds spatial
questions in designated regions and metric depth features~\cite{chen2024spatialvlm,cheng2024spatialrgpt}.
MM-Spatial studies open-category metric relations and the contribution of
depth and multi-view input~\cite{daxberger2025mmspatial}.  Video-3D LLM and
MV-ScanQA integrate multiple scene views using position-aware video features,
coverage-driven sampling, or 2D--3D--language pretraining~\cite{zheng2025video3dllm,mo2025mvscanqa}.
Their central target is recovering spatial relations or three-dimensional
scene information from accumulated observations.  Evaluating evidence
sufficiency further requires tracking the informative, repeated, or irrelevant
contribution of each view, together with the answers still permitted by blind
regions.

\subsection{Cross-View Identity}

Stable object identity is a prerequisite for combining views.  Classical
matching methods establish local correspondences through graph attention,
optimal transport, or detector-free dense matching~\cite{sarlin2020superglue,sun2021loftr}.
Re-identification and multi-camera tracking learn identity features across
viewpoint and camera changes~\cite{he2021transreid,ristani2018features,han2023mmptrack}.
MessyTable isolates instance association in cluttered tabletop scenes, where
repeated categories and occlusion make appearance-only matching
unreliable~\cite{cai2020messytable}.  Recent MLLM-oriented work reaches beyond
geometric keypoints: MMVM evaluates visual correspondence in general
multimodal models and trains object-level instance tokens, while ObjectRelator
transfers object conditions and masks between egocentric and exocentric
views~\cite{zhou2025mmvm,fu2025objectrelator}.  These methods can establish
that two observations refer to the same physical entity, which is essential
for avoiding duplicate counts.  Correspondence alone does not determine
whether an additional entity can remain outside the matched regions or behind
an occluder.

Object-centric representation learning provides a complementary mechanism for
organizing repeated evidence.  MONet, IODINE, and Slot Attention decompose a
scene into a set of object-like components through iterative inference or
competitive attention~\cite{burgess2019monet,greff2019iodine,locatello2020slotattention}.
Perceiver uses learned latent queries to compress large inputs, while DETR and
Mask2Former formulate detection and segmentation as query-based set
prediction~\cite{jaegle2021perceiver,carion2020detr,cheng2022mask2former}.
These models motivate compact entity sets.  Their components are normally
formed within one image or optimized for detection, segmentation, or scene
decomposition, leaving the evidential relation among arbitrary image subsets
unspecified.

Cross-view identity identifies repeated visual evidence.  Reasoning about a
claim additionally requires representing unobserved regions and the hidden
configurations they may still contain.

\subsection{Incomplete Evidence}

Natural language inference formalizes whether a premise entails, contradicts,
or remains neutral toward a hypothesis~\cite{bowman2015snli,williams2018multinli}.
Visual entailment transfers this three-way relation to image--text pairs, and
NLVR2 tests whether paired photographs satisfy compositional language
descriptions~\cite{xie2019visualentailment,suhr2019nlvr2}.  Diagnostic datasets
such as CLEVR and GQA further expose compositional and relational reasoning
failures through executable scene questions or structured scene
graphs~\cite{johnson2017clevr,hudson2019gqa}.  These benchmarks evaluate a
fixed image--text input or derive answers from a specified scene, without
certifying which alternative complete scenes remain consistent with the
visible evidence.  They consequently leave open how to evaluate the strongest
consequence shared by all complete worlds compatible with a partial
observation, including exact answers, bounds, and unresolved results.

Occlusion research also distinguishes visible evidence from the unobserved
extent of an object.  Amodal instance segmentation predicts complete object
shapes from visible fragments, with KINS providing large-scale annotations for
occluded driving scenes~\cite{li2016amodal,qi2019kins}.  For partial-observation
reasoning, a further question is whether an unobserved support region can
contain an additional instance and thereby change an existence or count claim.

An unresolved proposition also differs from unknown-category recognition and
model abstention.  Open-set recognition and open-world detection identify
inputs or object categories outside a registered label space~\cite{scheirer2013openset,joseph2021openworld}.
Selective classifiers reject predictions to control expected risk at a chosen
coverage level~\cite{geifman2017selectiveclassification,geifman2019selectivenet}.
Both address limitations of the predictor relative to its label space or
confidence estimate.  Their rejection behavior may change with recognition or
calibration quality even when the available evidence is unchanged.  A
different problem arises when the categories and proposition are known but
observationally equivalent complete worlds assign different answers.  Existing
abstention settings do not separate this evidence-induced indeterminacy from
model uncertainty, nor do they require weaker conclusions, such as a valid
lower bound, when an exact answer is unavailable.

VQA research has repeatedly shown that models can exploit question and answer
regularities instead of images~\cite{antol2015vqa,goyal2019making}.  Causal
counterfactual VQA removes a language-only direct effect from model scores,
while broader analyses describe such behavior as shortcut
learning~\cite{niu2021counterfactualvqa,geirhos2020shortcut}.  Such
counterfactuals diagnose language effects in a predictor.  Certifying
observational ambiguity requires physical examples of complete scenes with
identical observations and different hidden contents and answers.  Direct
supervision of these underdetermined conclusions also needs to control scene,
position, and language shortcuts.

\subsection{Partial Observability}

Partially observable decision processes summarize unobserved state with a
belief state and update it after each action and observation~\cite{kaelbling1998pomdp}.
Learned world models compress observations into latent dynamics for
future prediction and control~\cite{ha2018worldmodels,hafner2025dreamerv3}.
Both frameworks address sequential interaction: the former assumes action,
transition, observation, and reward models, while the latter emphasizes
future prediction or control.  They do not directly address the evidential
question of which conclusions already hold across all complete worlds
compatible with a given set of visual observations.

Active perception selects observations that reduce uncertainty or improve a
task-specific estimate~\cite{bajcsy1988activeperception}.
Learned active localization similarly frames visual information acquisition
as a sequential policy, although within a single image~\cite{caicedo2015activelocalization}.
These methods learn where to sense by executing actions in an image or
environment.  They leave open whether a model can assess the evidential effect
of already supplied observations or candidate images before an action is
executed.

\section{The \worldscope{} Dataset and Benchmark}
\label{sec:worldscope}

\subsection{Problem Formulation}

Let $W$ denote a complete tabletop world and let $S$ be a set of RGB
observations of that world.  Each observation exposes only part of the support
surface and a subset of its objects.  We study which conclusions $S$ supports.
A proposition $q$ is \emph{determined true} when every complete
world compatible with $S$ satisfies $q$, \emph{determined false} when none does,
and \emph{non-identifiable} when both truth values remain possible.  This
semantics also admits conclusions more informative than a three-way label.  An
observation may establish a count lower bound, rule out categories, or confirm
part of a regional distribution even when an exact answer remains unknown.

Formally, let $\Omega(S)$ be the set of complete worlds consistent with all
observations in $S$.  The evidential status of $q$ is
\begin{equation}
  \operatorname{status}(q,S)=
  \begin{cases}
    \mathrm{true}, & \forall W\in\Omega(S),\ q(W)=1,\\
    \mathrm{false}, & \forall W\in\Omega(S),\ q(W)=0,\\
    \mathrm{non\mbox{-}identifiable}, & \text{otherwise}.
  \end{cases}
  \label{eq:possible-world-status}
\end{equation}
Adding an observation can only contract this set:
$\Omega(S\cup\{v\})\subseteq\Omega(S)$.  A useful view therefore changes the
conclusions shared by the compatible worlds; an irrelevant or repeated view
can leave them unchanged.  This relation provides one semantics for single
images, unordered image sets, view removal, and sequential observation.

We instantiate this problem along three observation protocols.  \emph{Single
view} presents one partial observation, \emph{multi-view} provides a set of
observations to be composed, and \emph{active disambiguation} presents a base
observation together with candidate views or view orders.  Across the three
protocols, the underlying world property is drawn from eight families:
existence and absence, exact count, count thresholds and bounds, scoped count,
count comparison, category composition, regional distribution, and category
inventory.  Propositions combine existence, count, comparison, set, and
regional predicates with Boolean connectives.  Table~\ref{tab:worldscope-base}
summarizes these eight \emph{base query families}.  They specify what property
is queried.  The ten interfaces in Table~\ref{tab:worldscope-tasks} specify how
the available evidence is used or changed.  The two axes are composed whenever
their semantics are compatible: exact count, for example, can underlie a claim
judgment, a resolving-view question, a minimum sufficient set, or a direct
request for the tightest count interval.  Thus the eight base queries and ten
task interfaces are distinct, composable levels of the data design.

\begin{table}[t]
  \centering
  \renewcommand{\arraystretch}{1.04}
  \setlength{\tabcolsep}{2pt}
  \caption{Base-query taxonomy and data composition.}
  \label{tab:worldscope-base}
  \begin{tabularx}{\columnwidth}{@{}L{0.29\columnwidth}Zrr@{}}
    \toprule
    \textbf{Base query} & \textbf{Target} & \textbf{Training set} & \textbf{Benchmark} \\
    \midrule
    \rowcolor{wfopen}
    \multicolumn{2}{l}{\textbf{Presence and sets}} & \textbf{254.7K} & \textbf{1,197} \\
    Existence / absence & Confirmed presence / exclusion & 22.6K & 178 \\
    Category composition & Category-set membership & 30.9K & 214 \\
    Category inventory & Confirmed / possible categories & 201.2K & 805 \\
    \rowcolor{wfapi}
    \multicolumn{2}{l}{\textbf{Counting}} & \textbf{863.9K} & \textbf{3,402} \\
    Exact count & Unique count & 312.8K & 1,399 \\
    Count thresholds / bounds & Tightest supported count interval & 204.2K & 871 \\
    Scoped count & Count within a specified region & 222.8K & 662 \\
    Count comparison & Comparison between object counts & 124.0K & 470 \\
    \rowcolor{wfgreen}
    \multicolumn{2}{l}{\textbf{Regional reasoning}} & \textbf{81.4K} & \textbf{401} \\
    Regional distribution & Confirmed / possible regions & 81.4K & 401 \\
    \midrule
    \rowcolor{wfhuman}
    \textbf{Total} & & \textbf{1.2M} & \textbf{5,000} \\
    \bottomrule
  \end{tabularx}
\end{table}

\begin{table}[t]
  \centering
  \renewcommand{\arraystretch}{1.04}
  \setlength{\tabcolsep}{2pt}
  \caption{Task-interface taxonomy and data composition.}
  \label{tab:worldscope-tasks}
  \begin{tabularx}{\columnwidth}{@{}L{0.30\columnwidth}Zrr@{}}
    \toprule
    \textbf{Task} & \textbf{Operation} & \textbf{Training set} & \textbf{Benchmark} \\
    \midrule
    \rowcolor{wfopen}
    \multicolumn{2}{l}{\textbf{Claim reasoning}} & \textbf{360K} & \textbf{1,500} \\
    Claim status & Evaluate a claim from supplied views & 120K & 500 \\
    Claim selection & Compare candidate claims & 120K & 500 \\
    Strongest claim & Select the strongest supported claim & 120K & 500 \\
    \rowcolor{wfapi}
    \multicolumn{2}{l}{\textbf{Evidence operations}} & \textbf{480K} & \textbf{2,000} \\
    Claim-resolving view & Add a view to resolve a claim & 120K & 500 \\
    View necessity / redundancy & Remove each view and assess its role & 120K & 500 \\
    Shortest disambiguating sequence & Find the shortest resolving sequence & 120K & 500 \\
    Minimal sufficient view set & Compare subsets for sufficient evidence & 120K & 500 \\
    \rowcolor{wfgreen}
    \multicolumn{2}{l}{\textbf{Count acquisition}} & \textbf{240K} & \textbf{1,000} \\
    Best counting view & Add the view that maximizes confirmed count & 120K & 500 \\
    First count-gain view & Find the first view that increases the count & 120K & 500 \\
    \rowcolor{wfhuman}
    \multicolumn{2}{l}{\textbf{Maximal answer}} & \textbf{120K} & \textbf{500} \\
    Maximal identifiable answer & Return the strongest identifiable answer & 120K & 500 \\
    \midrule
    \rowcolor{wfhuman}
    \textbf{Total} & & \textbf{1.2M} & \textbf{5,000} \\
    \bottomrule
  \end{tabularx}
\end{table}

\subsection{Scene and Observation Construction}

We construct the main visual pool from 20 HSSD structural scenes spanning ten
indoor and semi-outdoor scene families~\cite{khanna2024hssd}.  For each
structure, we generate sparse, medium, and dense tabletop layouts containing
3, 6, and 9 target objects, respectively, with four layout instances per
density.  This produces 240 scene instances.  Target objects are sampled from
38 everyday tabletop categories and 153 object assets derived from
OmniObject3D~\cite{wu2023omniobject3d}.  The support surfaces and surrounding
rooms remain native to HSSD, with 19 support assets represented in the final
pool.  We add one to three larger contextual objects drawn from 18 categories
and 29 assets, including speakers, plants, monitors, and storage containers.
These objects create occluded regions within the image.  Category, asset,
placement region, scene family, density, and
occluder configuration are varied jointly so that hidden space is not tied to
a single visual pattern.

The construction follows a render-once, reason-many pipeline that separates
physical scene generation from question production.  A support surface is
selected in each HSSD room and populated with category-appropriate objects at
collision-free poses.  Layout generation checks support contact, boundary
containment, inter-object penetration, plausible scale, and visibility from at
least part of the camera set.  Occluders are then positioned to create locally
hidden objects and contiguous hidden surface regions.  Camera placement is
conditioned on the support geometry and surrounding room, preserving variation
in distance, elevation, perspective, and background while keeping the target
surface legible.  The resulting physical world is rendered once.  Offline
evidence extraction then measures per-view object visibility, cross-view
identity, and category-conditioned hidden space.  Executing query programs on
different image subsets yields structured answers, after which task interfaces
and language realizations are applied.  This order keeps rendering, evidence,
reasoning, and language as separately controlled stages.

Each scene is rendered from nine stable camera candidates using
Habitat-Sim~\cite{savva2019habitat}.  RGB observations have resolution
$1024\times768$
and are downsampled from $2\times$ supersampled renders.  Each question uses
the subset of camera views required by its observation protocol.  For
regional questions, a camera-matched reference image of the empty support
surface marks front-left, front-right, back-left, and back-right.  It defines
the coordinate convention but contributes no object or count evidence.

Rendering also produces aligned depth, semantic labels, and instance
identities.  These privileged modalities are used offline to certify
visibility, physical coverage, and cross-view correspondence.  They are never
part of a benchmark question or model input.  This separation provides exact
evidence labels from the simulator while preserving the intended inference
setting, in which a model receives only RGB images, their public roles, and the
question.  It also allows the same observation bank to support single-view,
multi-view, set, removal, and sequential tasks without changing the underlying
scene or its object identities.

The held-out pool uses five additional HSSD structures from five scene families
absent from the main pool.  Each structure contains two layout instances at
each density.  Two structures form an unseen-structure validation pool and
three form the final test set.  The partition is chosen to minimize divergence
in task and answer distributions without using model predictions.  Structural
scene identity is the split unit, preventing the same room geometry from
appearing in training and structure-disjoint evaluation.

\subsection{Evidence-Grounded Answer Generation}

\paragraph{Confirmed entities.}
Every physical object retains one identity across cameras.  An object is
confirmed only when its visible evidence exceeds the countability criterion;
small fragments do not become positive count evidence.  For category $c$,
region $r$, and observation set $S$, let $V_c(S,r)$ be the union of confirmed
object identities in $r$ over all views in $S$.  The confirmed lower bound is

\begin{equation}
  L_c(S,r)=\lvert V_c(S,r)\rvert .
  \label{eq:confirmed-lower-bound}
\end{equation}

Taking a union over identities is essential: the same object visible in three
images contributes once, whereas an object first exposed by a new image
increases the bound by one.

Countability is computed from each object's instance projection in each view.
This separates three cases
that would otherwise be conflated: an object absent from the world, an object
present but fully hidden, and an object whose visible fragment is too weak to
support reliable counting.  Once an object becomes countable in any selected
view, its stable identity enters $V_c(S,r)$ and remains confirmed for every
superset of $S$.

\paragraph{Object-size-aware hidden space.}
Absence and exact counts require more than enumerating visible objects.  We
densely sample the support surface and determine which locations are exposed by
the selected observations using aligned depth and semantic renderings.  A
remaining blind region is considered relevant to category $c$ only if an
object with that category's physical dimensions can be placed there without
leaving the support surface or intersecting known objects.  The placement test
uses a stable hidden neighborhood for small objects and permits a partially
visible feasible pose for large objects, reflecting that a small isolated gap
should not license an arbitrary hidden object while a large occluded volume can
still conceal most of a large object.  Visibility is checked jointly across
the selected observations.  Consequently, complementary views remove hidden
placements monotonically, while repeated views do not create spurious count
evidence.  Purely out-of-frame blind regions are retained when physically
necessary but are downweighted in question construction in favor of
occluder-caused blind regions.

Surface sampling adapts to the physical dimensions of the selected support.
Fine placement candidates
are evaluated over the actual support shape and several object orientations.
The test combines support boundaries, the footprints of visible objects, and
joint visibility across all observations.  It checks whether a physically valid
placement remains consistent with all supplied views.  Coverage is
category-dependent: a narrow blind
strip may remain open for a pen while being closed for a laptop.  The same
observation set can consequently establish an exact count for one category and
only a lower bound for another.

Let $H_c(S,r)$ indicate whether any feasible placement for another instance of
$c$ remains.  The observation-supported count interval is

\begin{equation}
\begin{aligned}
  I_c(S,r)&=[L_c(S,r),U_c(S,r)],\\
  U_c(S,r)=
  \begin{cases}
    L_c(S,r), & H_c(S,r)=0,\\
    \infty, & H_c(S,r)=1.
  \end{cases}
\end{aligned}
  \label{eq:count-interval}
\end{equation}

The infinity symbol denotes an open upper bound under the benchmark evidence
semantics; the generator never substitutes the privileged complete-scene count
when unseen space remains.  Existence, equality, threshold, comparison, set,
and regional propositions are evaluated over these intervals.  A proposition
is determined true if all compatible values satisfy it, determined false if no
compatible value does, and non-identifiable otherwise.  Maximal-identifiable-answer
examples report the maximal information licensed by the same evidence, including count
intervals, confirmed and excluded categories, and regions where an object is
confirmed or still possible.

\paragraph{Observation operations.}
Gold answers for view operations are obtained by executing the same evidence
calculation on the relevant image subsets.  A resolving view changes a claim
from non-identifiable to either true or false.  A view is necessary when
deleting it from the full set prevents the current judgment, and redundant
otherwise.  Minimal sets and shortest sequences are found by comparing all
admissible subsets or cumulative prefixes.  Count-view tasks compare the union
of confirmed identities before and after each addition.  Thus all ten task
interfaces share the same evidence semantics.

These computations form an evidence lattice over image subsets.  Nodes store
the strongest conclusion supported by one subset; edges correspond to adding
or removing one observation.  Set questions compare nodes at the same level,
while sequence questions follow paths through the lattice and locate the first
evidential change.  Because every node is recomputed from the same entity and
hidden-space semantics, the answer for a set is invariant to the order in
which its images are presented.  The lattice also supplies hard negative
examples: some views reveal additional pixels but no relevant entity or hidden
region, so their correct evidential effect is explicitly null.

\subsection{WorldScope-1.2M}

From the executable scene evidence, we construct 805,955 distinct reasoning
instances and realize them as 1.2 million English question--answer pairs over
4,281 unique RGB observations.  Each of the ten tasks contributes exactly
120,000 questions.  The eight base query families and ten task interfaces play
different roles in this count: a base query selects the property to infer,
whereas a task interface determines how an observation set is inspected,
expanded, reduced, or ordered.  The natural-language question is generated only
after its structured reasoning instance and answer have been fixed.

\begin{figure}[tp]
  \centering
  \begin{subfigure}{\linewidth}
    \centering
    \includegraphics[width=\linewidth]{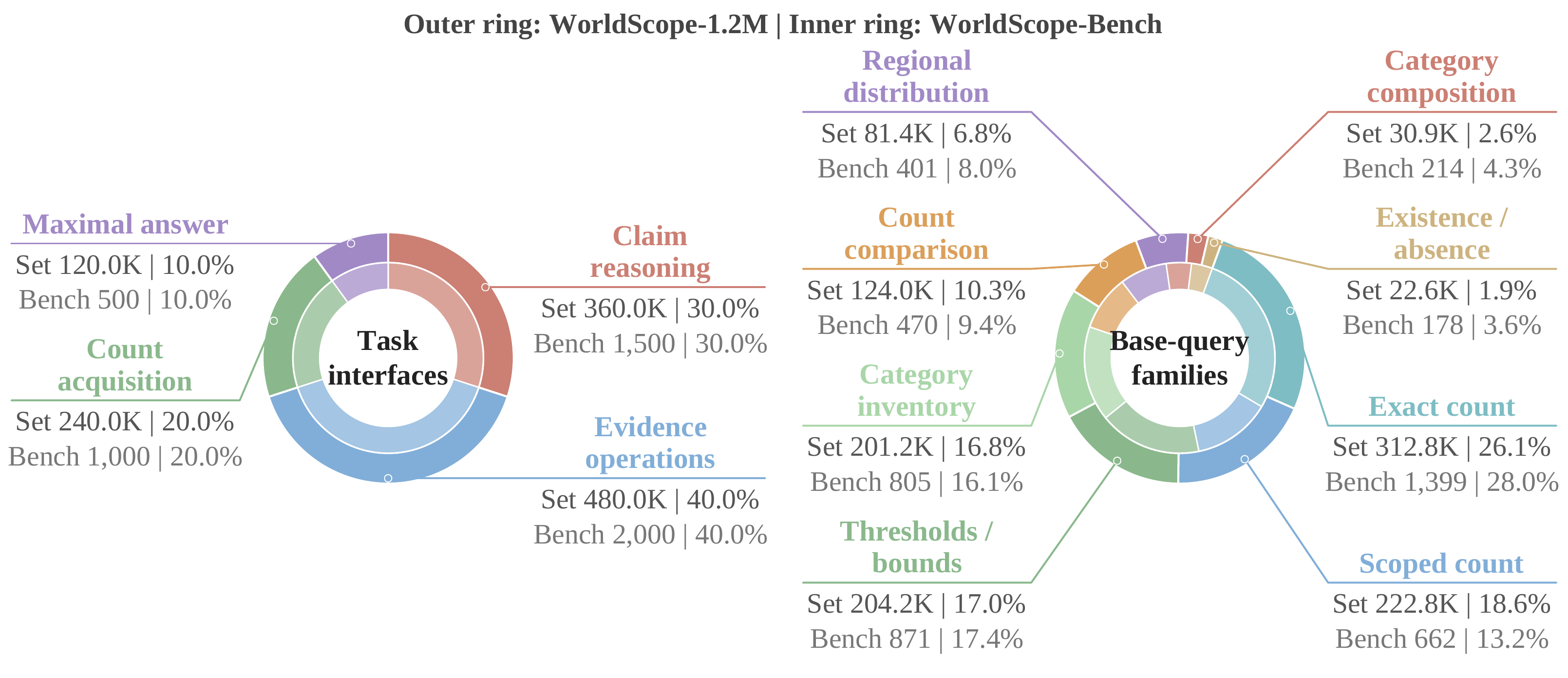}
    \caption{Training and benchmark composition.}
    \label{fig:dataset-composition}
  \end{subfigure}
  \par\medskip
  \begin{subfigure}{\linewidth}
    \centering
    \includegraphics[width=\linewidth]{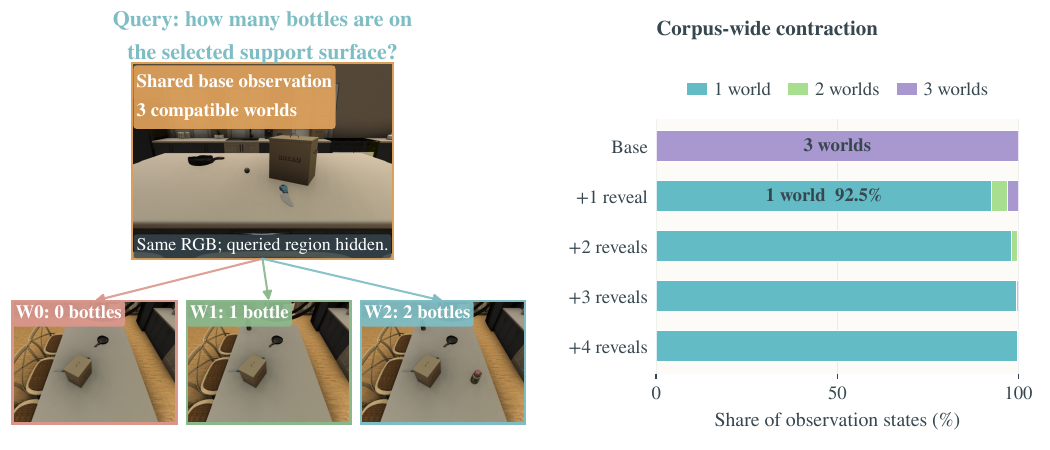}
    \caption{Certified counterworlds and evidence-induced contraction.}
    \label{fig:counterworld-certification}
  \end{subfigure}
  \caption{Data composition and certified counterworld supervision.
  (a) The two nested donut charts compare four task-interface groups (left)
  and eight base-query families (right). Outer rings denote \worldset{};
  inner rings denote \worldbench{}; callouts report counts and proportions.
  (b) A shared base observation admits three certified completions with zero,
  one, or two bottles; a reveal view exposes their differences. The chart
  summarizes compatible-set contraction over 446,400 image-subset states from
  4,800 groups. These completions certify ambiguity but do not exhaust all
  physically possible worlds.}
  \Description{Two subfigures. The upper subfigure contains horizontally
  aligned composition donuts. The lower subfigure places a shared image and
  three reveal images beside a stacked bar chart of compatible-world counts.}
  \label{fig:dataset-overview}
\end{figure}

Figure~\ref{fig:dataset-composition} compares the training and benchmark
composition along both axes of the data design.

Candidate generation begins from the full executable lattice of each scene.
We enumerate valid propositions from the eight base query families, evaluate
them on single views and admissible multi-view subsets, and derive task records
only when the requested contrast is well-defined.  Resolving-view examples, for
instance, require an unresolved base observation and at least one candidate
that reaches a definite truth value; shortest-sequence examples require a
resolving prefix whose length can be compared across candidate orders.  This
eligibility filtering prevents questions whose answer is determined by a
template convention rather than by visual evidence.  The structured answer is
computed before the same reasoning instance is verbalized as a question.

The final sample is selected from a substantially larger candidate pool.
Sampling is stratified over task, base query, truth status or answer-set shape,
scene structure, density, category, number of supplied views, and the location
of the informative view.  Per-scene caps prevent a small number of productive
layouts from dominating the corpus, and rare-category reweighting retains the
long tail of the object inventory.  Option labels are assigned after the
evidential answer is fixed, and image display order is independently permuted.
For selection and sequence tasks, positive-set cardinality and answer position
are balanced separately.  These steps weaken question-only, camera-only,
fixed-position, and generation-order correlations without changing the
underlying evidence semantics.  The resulting density distribution contains
318,593 sparse, 393,559 medium, and 487,848 dense questions.

\subsection{Certified Counterworld Supervision}

We complement the interval-based language targets with visual examples of
observational ambiguity.  We construct 4,800 training-only counterworld
groups over all 20 main structures and all 240 scene instances.  Candidate
groups start from a common visible scene and a support region hidden from a
chosen base camera.  We instantiate three physically valid completions by
changing the category-consistent hidden objects and their collision-free poses
while preserving everything observable from the base camera.  Each completion
is then rendered from the shared base camera and five reveal cameras.

The base observations are accepted only when the three completions agree in
RGB, depth, and semantic renderings, which rules out imperceptible color
differences as well as geometric or categorical leakage.  At least one reveal
view must make some completion distinguishable, and the three complete worlds
must yield at least two answers to the associated query.  Among the accepted
groups, 3,491 contain two distinct completion answers and 1,309 contain three.
Thus every shared base image has a concrete pair of physically rendered worlds
that agree on the observation but disagree on the answer.

We next compare each of the six observations rendered from every completion
against each of the three complete-world hypotheses, producing 259,200
view--world compatibility judgments.  These atomic judgments are composed over
every admissible reveal subset: a completion remains compatible only when every
supplied observation is consistent with it.
For each subset we record the compatible completions, and for every edge formed
by adding one view we record which completions are eliminated.  This produces
446,400 image-subset compatibility states and 1,080,000 one-view transitions
from 14,400 complete worlds.  Uninformative edges preserve the compatible set;
revealing edges contract it.

Figure~\ref{fig:counterworld-certification} connects this supervision at the
instance and corpus levels.  In the illustrated group, the shared observation
hides the queried region, while a reveal view separates three physically valid
completions with different bottle counts.  Across all groups, the base
observation retains all three certified worlds, whereas one reveal view leaves
a single compatible world for 92.5\% of the resulting observation states.

The three completions are concrete counterexamples rather than an exhaustive
enumeration of all possible worlds.  Disagreement between two compatible
completions certifies non-identifiability, while agreement among the sampled
completions alone is not treated as proof that all possible worlds agree.  The
4,800 groups are selected with source-frequency weighting over structure,
scene, density, category, camera, placement region, and object asset.  They
cover all 38 target categories and all 153 target assets and broadly follow the
visual distribution of \worldset{}.  The counterworld corpus contributes no
synthetic propositions or question--answer pairs.  All language supervision
continues to come from \worldset{}, so the auxiliary world supervision does not
introduce a second question distribution.

\subsection{WorldScope-Bench}

\worldbench{} is generated with the same evidence executor, eight base query
families, ten task interfaces, and three observation protocols as \worldset{},
but from five additional HSSD structures belonging to scene families absent
from the training pool.  Candidate construction first enumerates all valid
propositions and observation operations in these scenes, then keeps only items
whose answer is visually grounded and whose requested contrast exists.  This
matters for set and sequence tasks: an image set must have a genuine sufficient
subset, and a proposed order must contain a measurable first resolving step.
Questions that can only be answered by a template convention are excluded.

The five structures are partitioned at the structure level.  We evaluate all
ten choices of two development and three test structures and choose the split
with the smallest combined Jensen--Shannon divergence in task and answer
distributions.  No model prediction is used in this choice.  Development and
test sets also contain questions sampled from the 20 main structures, allowing
performance under familiar geometry to be separated from structural transfer.
For every task, development contains 200 seen-structure and 200 unseen-structure
questions, giving 4,000 questions in total.  Test contains 200 seen-structure
and 300 unseen-structure questions per task, giving 5,000 questions in total.
The 3,000 unseen test questions come from 18 scene instances in the three
structure-disjoint test structures.  Balanced task quotas make both the task
average and the overall question accuracy directly interpretable.

Official scoring uses structured answers; optional free-form explanations do
not affect the score.  We compute exact accuracy separately for every task and
macro-average the ten values.  Since all tasks have equal quotas, this macro
average equals overall question accuracy.  Single-choice and status tasks use exact match.
Multi-answer proposition and view tasks also report set F1, while exact credit
requires the complete set with neither missing nor spurious choices.
Necessity/redundancy requires both partitions to be correct.  A shortest-sequence
prediction is accepted when it matches any gold sequence attaining the minimum
resolving-prefix length.  For maximal identifiable answers, the lower bound,
upper bound, exactness flag, category sets, and region sets are scored as
components, with exact credit granted only when every applicable component is
correct.  The metric therefore tests whether a model returns the strongest
claim licensed by the supplied observations and tracks how that claim changes
under view addition, removal, grouping, and order.

\subsection{Quality Control and Human Review}

Quality control is applied at the scene, evidence, and question levels.
Automatic checks reject invalid support placement, object penetration,
implausible scale, corrupted rendering, inconsistent cross-view identity, and
any evidence transition that violates monotonicity.  We manually review every
rendered scene for physical plausibility, recognizable assets, meaningful
occlusion, and usable camera framing, removing or repairing scenes that fail
these criteria.

For \worldset{}, we draw a stratified random sample of 12,000 training questions
across tasks, base queries, answer forms, densities, and scene families.  Each
sample is checked for image--question correspondence, the sufficiency of visible
evidence, language clarity, and the structured gold answer.  Every candidate
question for \worldbench{} is then reviewed individually.  We remove items with
physically implausible hidden placements, ambiguous visual evidence, confusing
wording, or an incorrect answer contract.  After this full benchmark review,
97.2\% of the candidates are retained in the released evaluation pool.

\section{Methodology}
\label{sec:method}

\subsection{Overview}

We propose \textbf{WorldFlow}, an evidence adapter for visual reasoning under
partial observation.  WorldFlow represents every image subset from two
complementary sides.  \emph{Observed evidence} summarizes the entities that
the selected images jointly confirm.  \emph{Latent evidence} summarizes the
physical support regions and complete-world configurations that those images
have not yet ruled out.  The first side establishes visible facts and count
lower bounds; the second prevents these facts from being mistaken for a
complete description of the world.

For a sample with $n\leq5$ images, Qwen's vision encoder runs once on each RGB
image.  WorldFlow compresses the resulting patch features into 64 observation
tokens and enumerates the $2^n-1$ non-empty image subsets.  For every subset,
16 cross-view entity slots form its observed evidence, while a $4\times4$
support-surface coverage map forms its latent evidence.  A shared encoder
combines both sides with the selected image features and produces four subset
tokens.  All subset representations together form an \emph{evidence lattice}:
each node corresponds to one image set, and adjacent nodes differ by one
observation.

The subset tokens are projected into Qwen's language space and inserted
alongside the original image tokens and question.  This information flow can be
summarized as \emph{observe, compose, constrain, and conclude}: encode each
observation, compose its two-sided evidence, represent how an image set
constrains the world, and generate the strongest conclusion licensed by those
constraints.  These terms describe successive operations within a single
forward computation.  The same representation serves every task interface;
the model receives no task identifier and uses no task-specific answer head.

\begin{figure}[t]
  \centering
  \includegraphics[width=\textwidth]{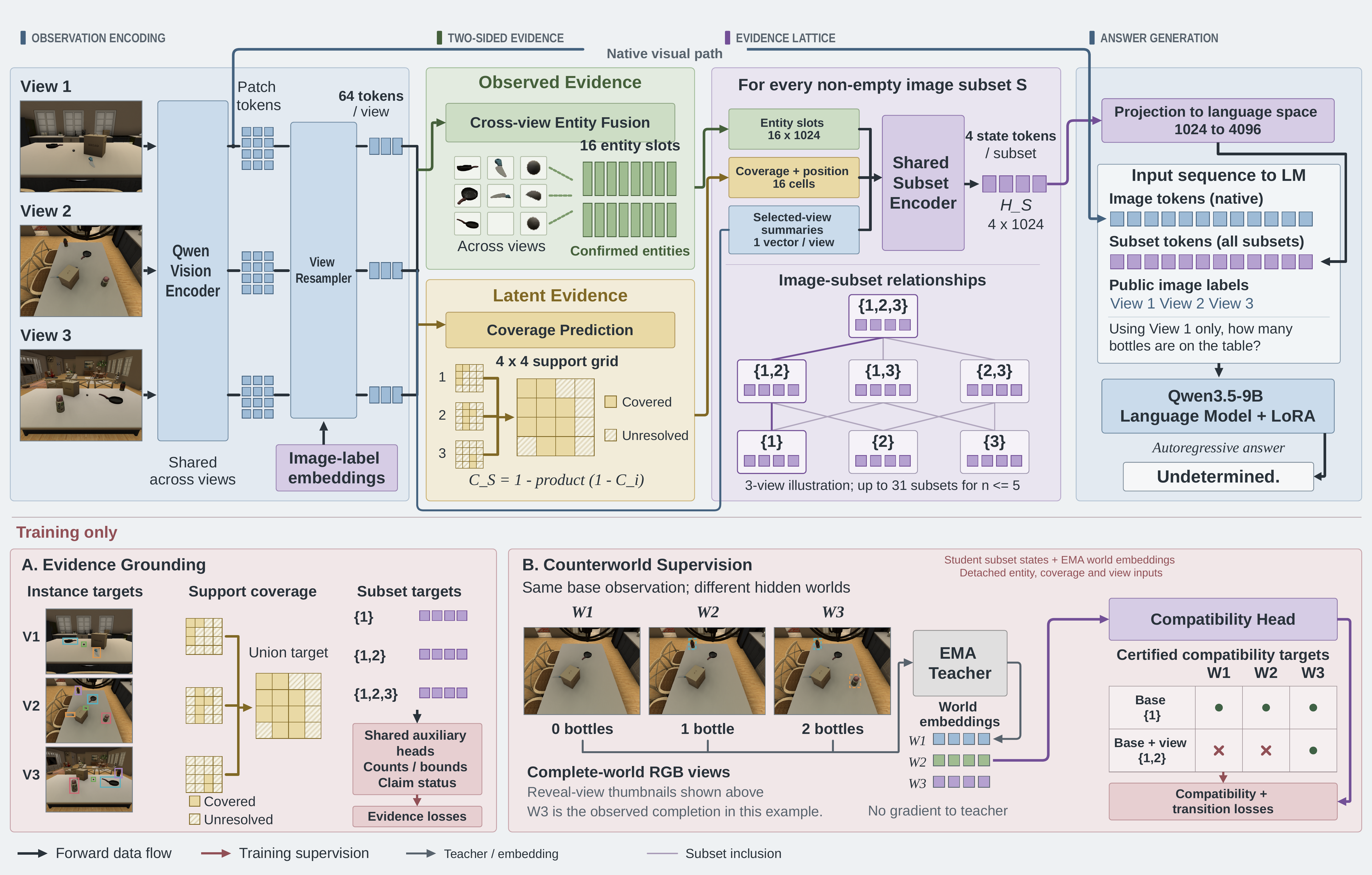}
  \caption{WorldFlow architecture. Each RGB view is compressed by the shared
  Qwen vision encoder and resampler. Observed entity slots and latent support
  coverage are composed for every non-empty image subset, whose four state
  tokens form an evidence lattice before projection into Qwen's language
  space. The original image tokens remain in the multimodal sequence. The EMA
  teacher, complete-world embeddings, and compatibility head are used only for
  training; the coverage grids, token arrays, and counterworld thumbnails are
  schematic architectural illustrations rather than checkpoint outputs.}
  \Description{A full-width architecture diagram showing per-view encoding,
  two-sided evidence branches, subset-state lattice, answer generation, and
  training-only counterworld compatibility supervision.}
  \label{fig:worldflow-architecture}
\end{figure}

\subsection{Observation Encoding}

Let $X_i\in\mathbb{R}^{P_i\times4096}$ denote the Qwen patch features of image
$i$, and let $M_i$ mark valid patches.  A learned projection maps $X_i$ into a
1024-dimensional evidence space.  A resampler with 64 learned queries then
reads the projected patches through multi-head attention:

\begin{equation}
  E_i=\operatorname{LN}\left(Q+\operatorname{MHA}(Q,\,X_iW_x,\,X_iW_x;M_i)\right),
  \qquad E_i\in\mathbb{R}^{64\times1024}.
  \label{eq:view-resampler}
\end{equation}

The public image label is encoded with Qwen's input embedding, projected to the
same space, and added to every token in $E_i$.  It preserves the association
between images and roles such as base view or candidate view, but contains no
task number or teacher annotation.  The same resampler is used for every image
and task.

\subsection{Observed Evidence}

For subset $S$, the observed-evidence module reads only the 64-token sequences
of its member images.  Sixteen learned slot queries jointly attend to their
concatenation, followed by a residual feed-forward block:

\begin{equation}
  Z_S=\operatorname{LN}\left(\operatorname{MHA}(Q_e,\,E_S,\,E_S)+Q_e\right),
  \qquad Z_S\in\mathbb{R}^{16\times1024}.
  \label{eq:entity-slots}
\end{equation}

Each slot denotes one stable entity candidate across the selected views.  A
slot is paired with every member-view summary to predict its instance mask on
the Qwen patch grid, countability logit, and normalized box in that view; the
slot itself also predicts an occupancy logit.  The mask query combines the
shared slot with a view summary and scores it against projected raw patch
features using normalized dot products.  One entity can therefore keep a
shared subset representation while changing location and appearance across
cameras.  Each slot also produces an open-vocabulary category embedding.

Slot indices are not treated as permanent object IDs.  For each subset,
Hungarian matching assigns slots to stable-object targets jointly over all
member views.  Its cost combines mask Dice and binary cross-entropy,
countability, box L1, and category-embedding distance.  A
stable object is matched to one slot, so its repeated appearance does not
increase the confirmed count.  For nested subsets, a consistency loss brings
the matched representations of already-countable objects closer.  An object
first revealed by a new image is free to occupy a previously empty slot.  The
slots and their countability predictions consequently describe the distinct
entities confirmed by the current observations and support a count lower
bound.  They cannot, by themselves, exclude an additional object in an unseen
region.

\subsection{Latent Evidence}

The latent-evidence branch summarizes the part of the support surface that
remains unresolved.  It mean-pools the 64 tokens of each image and predicts 16
sigmoid values, one for each cell of a $4\times4$ grid.  Coverage for subset
$S$ is composed by a probabilistic union:

\begin{equation}
  C_S[x]=1-\prod_{i\in S}(1-C_i[x]),
  \qquad x\in\{1,\ldots,16\}.
  \label{eq:coverage-union}
\end{equation}

A second head maps $C_S$ to a coverage-complete logit.  The grid is not an
image-space visibility map.  It estimates how much of the physical support
surface has been ruled out as unobserved placement space.  Its dense,
category-neutral target is generated offline from the support surface, and
cells with no legal support locations are excluded by a validity mask.  The
category- and size-dependent placement tests used by the data executor provide
the physical basis for this supervision, while depth, semantic masks, legal
sites, and stable IDs never enter the model input.

Coverage gives a local description of unexcluded surface space.  During
training, certified counterworlds add a global description of what remains
possible: they show which distinct complete configurations are still
compatible with the current RGB observations.  This global signal shapes the
subset representation above the local coverage branch.  Counterworld representations
are used only by the learning objective described in
Section~\ref{sec:method-learning}; they are not available during inference.

\subsection{Evidence Lattice}

WorldFlow combines the observed and latent sides separately for every non-empty
image subset.  The subset encoder concatenates 16 entity slots, 16 projected
coverage cells with learned grid-position embeddings, and one projected image
summary for every selected view.  Four learned queries, conditioned on the
mean summary of the subset members, attend to this source:

\begin{equation}
  \begin{aligned}
  T_S&=[Z_S;G_S;R_S],\\
  A_S&=\operatorname{MHA}(Q_s+g(S),T_S,T_S),\\
  H_S&=\operatorname{LN}(A_S+Q_s+g(S)),
  \end{aligned}
  \qquad H_S\in\mathbb{R}^{4\times1024}.
  \label{eq:subset-state}
\end{equation}

Here $G_S$ denotes the positional coverage representation, $R_S$ the selected
image summaries, and $g(S)$ the projected member summary.  The four tokens in
$H_S$ summarize the evidence available from $S$.  They are not four
possible-world slots and do not enumerate complete scenes.  Their mean feeds
shared auxiliary heads for confirmed object count (17 classes, 0--16), count
lower bound (17 classes), finite upper bound (binary), and claim status
(determined true, determined false, or non-identifiable).  For these auxiliary
predictions, a mean-pooled embedding of the natural-language question is
projected into the same space and added to the pooled state.

All lattice nodes use the same encoders and differ only in their selected
images.  A single-view or ordinary multi-view question reads the corresponding
node; view addition and removal compare nodes connected by one observation;
set questions compare several nodes; and sequence questions follow cumulative
subsets through the lattice.  Thus the ten task interfaces operate on one
shared representation.

\subsection{Answer Generation}

The wrapper builds Qwen's ordinary multimodal chat sequence by interleaving
each public image label with its RGB image and appending the original question.
For $n$ images, it additionally places $(2^n-1)\times4$ reserved state
placeholders between the images and question.  Each $H_S$ is projected from
1024 to Qwen's 4096-dimensional language space and written into its
corresponding positions.  Qwen's native image features are inserted in
parallel through the standard visual path, after which the multimodal position
indices are recomputed.  The language model therefore receives both the subset
tokens and the original image features.

All tasks use Qwen's shared autoregressive output layer and retain their native
answer strings.  At inference, deterministic generation receives only the RGB
images, public image labels, and question.  The complete-world teacher,
compatibility head, answer, scene metadata, and privileged sensor arrays are
absent.

\subsection{Learning}
\label{sec:method-learning}

Training follows a three-stage curriculum before the final model is used for
evaluation.  The first stage runs 10,000 scene-supervision updates.  It trains
the WorldFlow evidence modules with entity, coverage, and image-subset targets
while the Qwen backbone remains frozen.  This stage grounds the representation
in observable entities, countable visibility, and uncovered support regions.

The second stage runs 4,000 world-compatibility updates from the first-stage
checkpoint.  The entity and coverage branches are kept fixed; the subset-state
encoder and the training-only compatibility head learn from certified complete
worlds and from the worlds eliminated by an additional view.  The teacher
encoder is updated as an exponential moving average during this stage.  This
separates the learning of global observation compatibility from the grounding
of local visual evidence.

The third stage runs 4,000 joint updates.  Each eight-update cycle contains six
QA updates, one scene update, and one compatibility update.  QA updates train
Qwen with the answer-token causal loss and LoRA parameters, while the scene and
compatibility updates retain the auxiliary evidence objectives.  The later
joint-training configuration may unfreeze the last four Qwen vision layers;
this parameter change takes place within the joint stage.  Across all stages,
scene records provide entity, coverage, and subset
targets; compatibility records provide certified complete worlds and one-view
transitions; QA records provide the answer string.  Visual batches always retain
all single-image subsets and the full set, while intermediate subsets can be
rotated to bound the cost of computing up to 31 nodes.

\paragraph{Evidence grounding.}
Matched entity slots use mask binary cross-entropy and Dice losses, box L1 plus
generalized-IoU loss, countability and occupancy binary cross-entropy, and
open-vocabulary category alignment.  Their weights are 1.0, 1.0, 0.25, 0.5,
0.5, and 0.1, respectively; nested-subset slot consistency has weight 0.1.
Coverage-grid and coverage-complete losses each have weight 0.5, with the
complete loss multiplied by three for incomplete targets.  Confirmed-count and
lower-bound cross-entropy have weights 0.5 and 0.25, finite-upper-bound binary
cross-entropy has weight 0.25, and claim-status cross-entropy has weight 0.5.
Together these targets connect the two-sided subset representation to confirmed
entities, count bounds, and remaining support space.

\paragraph{Possibility learning.}
For each certified group, a frozen teacher encodes the RGB views of every
complete world.  The teacher is an exponential-moving-average copy of the
student WorldFlow core with momentum 0.995.  Completions with more than five
views are encoded in chunks and combined according to their represented view
counts.  A training-only compatibility head compares a student subset state
with each completion using projected cosine similarity and learned pair
functions over both representations, their absolute difference, and their
elementwise product.  The head predicts an independent compatibility logit for
each world.

The compatibility and in-group ranking losses each have weight 0.2.  A
transition loss with weight 0.2 requires a world eliminated by a new view to
cross the incompatibility boundary; a preservation loss with weight 0.05 keeps
scores stable for uninformative views.  Entity and coverage inputs are detached
within this branch, so counterworld supervision shapes the subset encoder
without overwriting grounded perception.  The teacher and compatibility head
are discarded after training.

\paragraph{Answer learning.}
The Qwen chat template appends the gold answer, while all labels before the
assistant response are masked.  The causal language loss is therefore computed
only on answer tokens and has weight 1.0.  Per-example losses are scaled by
source-distribution weights, and full-model QA batches add claim-status
supervision with coefficient 0.5.  Qwen's language backbone is adapted with
LoRA, while the WorldFlow modules are trained directly.  Scene and
counterworld annotations supervise the auxiliary objectives.  In both training
and inference, the student receives RGB images, image labels, and the question.

\section{Experiments}
\label{sec:experiments}

\subsection{Experimental Setup}

\paragraph{Model and optimization.}
All \worldflow{} variants and local ablations use the Qwen3.5-9B
backbone~\cite{qwen2026qwen35}, the same RGB images and public image labels,
and the same question and answer contracts.  The full curriculum comprises
10,000 scene-supervised updates, 4,000 counterworld-compatibility updates, and
4,000 joint updates.  New modules use a learning rate of $2\times10^{-4}$ in
the first two stages and $5\times10^{-5}$ in the joint stage.  Rank-64 LoRA
with scaling 128 and dropout 0.05 adapts the language model at $10^{-5}$.
AdamW uses weight decay 0.1, 3\% warmup followed by cosine decay, gradient
clipping at 1.0, BF16, and gradient checkpointing.  Effective batch sizes are
126 in the two warmup stages and 128 in joint training.  The reported model
keeps the Qwen vision backbone frozen and directly optimizes the \worldflow{}
modules.

\paragraph{Splits and metrics.}
The development set has 4,000 questions: each task contributes 200 questions
from seen structures and 200 from two unseen structures.  It is used for
checkpoint selection.  The test set has 5,000 questions: each task contributes
200 seen-structure and 300 unseen-structure questions.  The 3,000 unseen
questions come from three HSSD structures excluded from training and model
selection.  We report exact accuracy for every task and its average over all
ten tasks.  Since every task contributes 500 test questions, macro accuracy
and overall question accuracy are identical.  Multi-answer tasks require the
complete predicted set, and maximal identifiable answers require all applicable
fields to be correct.

\paragraph{Comparisons and ablations.}
The main comparison includes native Qwen3.5-9B, the same backbone trained only
on \worldset{} QA, and general-purpose multimodal models.  Models not trained
on \worldscope{} use a permissive parser that accepts semantically equivalent
answer formats.  Architecture ablations add entity evidence, coverage evidence,
the evidence lattice, and counterworld supervision cumulatively, then remove
the entity or coverage branch from the full model.  Curriculum ablations keep
the full architecture and remove Stage~1, both warmups, auxiliary replay in
Stage~3, or Stage~2.  Every table reports absolute accuracy.

\begin{figure}[t]
  \centering
  \includegraphics[width=\textwidth]{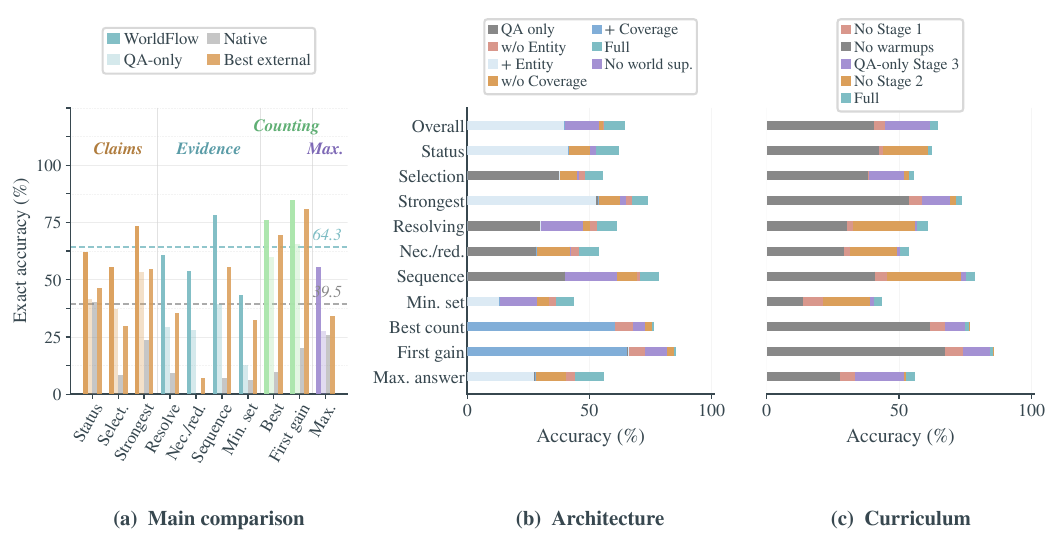}
  \caption{Task-wise exact accuracy on the complete 5,000-question test set.
  (a) WorldFlow versus the same-backbone QA-only and native models and the
  strongest external result on each task.  (b) Architecture variants and
  (c) curriculum variants.  WorldFlow improves over QA-only on all ten tasks;
  the ablations localize the gains to evidence composition and its supervision.}
  \Description{Three task-wise accuracy panels.  The left panel uses vertical
  grouped bars to compare WorldFlow with QA-only, native Qwen, and the best
  external model on ten tasks.  The two right panels compare architecture and
  curriculum ablations on the same tasks.}
  \label{fig:result-panels}
\end{figure}

\subsection{Main Comparison}

Table~\ref{tab:main-comparison} reports the complete 5,000-question test.  Full
\worldflow{} reaches 64.34\%, compared with 39.46\% for the same backbone
trained on QA alone and 15.06\% for native Qwen3.5-9B.  The 24.88-point
same-backbone gain is consistent across all ten tasks.  \worldflow{} also
reaches 50.43\% on the 3,000 structure-disjoint questions, compared with
35.60\% for QA-only and 13.63\% for the native backbone.

Figure~\ref{fig:result-panels}(a) gives the task-wise comparison against QA-only,
native Qwen3.5-9B, and the strongest external result obtained on each task.

\begin{table}[t]
  \centering
  \caption{Main comparison on the balanced 5,000-question test set.  Values
  are exact accuracy (\%); models are sorted by overall accuracy.  The best
  result in each column is shown in bold red. Task headings identify claim
  reasoning, evidence operations, count acquisition, and maximal answers.}
  \label{tab:main-comparison}
  \renewcommand{\arraystretch}{1.03}
  \setlength{\tabcolsep}{0.5pt}
  \fontsize{9}{11}\selectfont
  \begin{tabularx}{\textwidth}{@{}>{\raggedright\arraybackslash}p{0.16\textwidth}*{11}{Y}@{}}
    \toprule
    \rowcolor{wfheader}
    \textbf{Model}
      & \wfmainhead{\textbf{Claim}\\\textbf{status}} & \wfmainhead{\textbf{Claim}\\\textbf{select.}} & \wfmainhead{\textbf{Strongest}\\\textbf{claim}} & \wfmainhead{\textbf{Resolve}\\\textbf{view}}
      & \wfmainhead{\textbf{Necess./}\\\textbf{red.}} & \wfmainhead{\textbf{Shortest}\\\textbf{seq.}} & \wfmainhead{\textbf{Minimal}\\\textbf{set}} & \wfmainhead{\textbf{Best}\\\textbf{count}}
      & \wfmainhead{\textbf{First}\\\textbf{gain}} & \wfmainhead{\textbf{Max.}\\\textbf{answer}} & \textbf{Overall} \\
    \midrule
    \rowcolor{wfgreen}
    \shortstack[l]{\textbf{\worldflow{}}\\\textbf{(Qwen3.5-9B)}} & \best{62.00} & \best{55.40} & \best{73.60} & \best{60.80} & \best{53.60} & \best{78.20} & \best{43.20} & \best{76.00} & \best{85.00} & \best{55.60} & \best{64.34} \\
    \rowcolor{wfblue}
    Qwen3.5-9B\textsuperscript{a} & 41.40 & 37.20 & 53.20 & 29.40 & 28.00 & 39.60 & 12.80 & 60.00 & 65.40 & 27.60 & 39.46 \\
    \rowcolor{wfopen}
    Gemini 3.1 Pro & 40.40 & 30.00 & 54.80 & 32.80 & 3.60 & 38.40 & 20.80 & 55.20 & 74.00 & 24.40 & 37.44 \\
    Gemini 3 Pro & 45.60 & 29.20 & 39.60 & 32.60 & 3.40 & 44.00 & 32.20 & 36.20 & 78.40 & 30.60 & 37.18 \\
    GPT-5.6-Luna & 33.00 & 25.60 & 44.20 & 23.20 & 7.00 & 47.00 & 18.20 & 69.40 & 72.40 & 30.20 & 37.02 \\
    GPT-5.6-Terra & 42.80 & 26.40 & 47.20 & 25.60 & 5.60 & 38.00 & 19.20 & 50.20 & 81.00 & 31.80 & 36.78 \\
    GPT-5.6-Sol & 33.20 & 23.80 & 28.00 & 35.40 & 4.80 & 49.20 & 14.80 & 50.00 & 68.00 & 28.40 & 33.56 \\
    Claude Opus 4.8 & 35.60 & 7.60 & 23.40 & 11.80 & 2.40 & 55.60 & 7.80 & 30.80 & 51.00 & 29.80 & 25.58 \\
    Claude Opus 5 & 30.80 & 9.80 & 30.80 & 8.80 & 2.60 & 55.20 & 8.00 & 30.60 & 40.60 & 26.20 & 24.34 \\
    Grok 4.5 & 45.20 & 12.20 & 54.20 & 12.00 & 3.80 & 23.20 & 2.60 & 11.60 & 41.00 & 33.60 & 23.94 \\
    Qwen3.7-Max & 45.40 & 22.80 & 18.60 & 23.80 & 6.00 & 34.80 & 12.60 & 12.00 & 44.80 & 11.00 & 23.18 \\
    Qwen3.8-Max & 46.40 & 17.60 & 24.00 & 14.20 & 6.40 & 38.20 & 4.20 & 12.40 & 53.00 & 12.00 & 22.84 \\
    Claude Opus 4.7 & 30.60 & 16.60 & 31.40 & 8.40 & 2.20 & 41.40 & 7.40 & 25.80 & 36.20 & 27.60 & 22.76 \\
    Kimi-K2.6 & 36.80 & 12.60 & 43.60 & 9.20 & 4.20 & 29.40 & 3.00 & 31.20 & 24.80 & 31.00 & 22.58 \\
    Qwen3.7-Plus & 45.80 & 22.40 & 19.40 & 14.00 & 5.40 & 35.00 & 15.20 & 11.20 & 39.60 & 10.00 & 21.80 \\
    Qwen3.6-Plus & 44.60 & 23.60 & 29.00 & 10.40 & 5.00 & 39.60 & 3.60 & 10.80 & 40.80 & 9.00 & 21.64 \\
    Qwen3-VL-8B & 31.20 & 8.40 & 31.00 & 26.00 & 5.20 & 15.40 & 3.20 & 10.40 & 50.00 & 34.00 & 21.48 \\
    Grok 4.6 & 29.60 & 7.80 & 21.60 & 10.00 & 4.40 & 39.20 & 2.80 & 16.60 & 46.20 & 29.40 & 20.76 \\
    \rowcolor{wfhuman}
    Qwen3.5-9B\textsuperscript{b} & 40.20 & 8.20 & 23.80 & 9.40 & 0.00 & 7.00 & 6.20 & 9.80 & 20.00 & 26.00 & 15.06 \\
    \bottomrule
  \end{tabularx}
  \par\smallskip
  {\small\raggedright \textsuperscript{a}QA-only fine-tuning on \worldset{}. \textsuperscript{b}Native backbone without \worldscope{} training. Gemini Pro rows use Preview checkpoints; Qwen3-VL-8B denotes the Instruct checkpoint. Task names follow Table~\ref{tab:worldscope-tasks}; Necess./red. denotes view necessity and redundancy.\par}
\end{table}

The largest same-backbone gains occur on shortest sequence (+38.60), resolving
view (+31.40), minimal sufficient set (+30.40), and view necessity and
redundancy (+25.60).  These tasks compare image subsets or evidence transitions and
therefore benefit most from the explicit subset representation.  Native
Qwen3.5-9B obtains 0\% on necessity/redundancy because, under the 1,024-token
generation budget, none of its responses could be parsed into both required
sets.  The task is unusually sensitive to answer formatting even with
permissive parsing, so this value should not be read as an isolated measure of
visual ability.

\subsection{Architecture Ablations}

Table~\ref{tab:arch-ablation} separates the cumulative construction from the
two leave-one-out tests.  The +Entity and +Coverage variants remain close to the
QA-only baseline on unseen structures (35.57\% and 35.70\%, compared with
35.60\%), with only small finite-sample fluctuations because their outputs are
not connected to answer generation without the evidence lattice.  The first
substantial architectural gain appears when the lattice makes image subsets
directly comparable, raising unseen accuracy to 45.97\%; counterworld
supervision then raises it further to 50.43\%.  Removing entity evidence from
the full model costs 7.03 points; removing coverage costs 8.00.

Figure~\ref{fig:result-panels}(b) shows the corresponding task-wise results
on the combined test set.

The task pattern matches the intended division of labor.  Removing entity
evidence causes the largest unseen drop on first count gain (14.00 points),
where duplicate instances must be merged across an observation sequence.
Removing coverage has little effect on the two count-view tasks, but reduces
strongest claim, resolving view, necessity/redundancy, and maximal identifiable
answer by 10.67--12.00 points.  The evidence lattice is most visible on
resolving view, shortest sequence, and minimal sufficient set, each of which
must compare several image subsets.  Counterworld supervision adds a further
8.66 points on resolving view and 9.00 on minimal sufficient set beyond the
lattice-only model.

\subsection{Training Curriculum Ablations}

Table~\ref{tab:curriculum-ablation} tests how the same architecture changes
with its curriculum.  Removing Stage~1 lowers unseen accuracy by 11.36 points;
starting without either warmup lowers it by 14.23.  Keeping both warmups but
using only QA updates in Stage~3 is much closer to the full model, with a
2.90-point deficit.  Removing only Stage~2 gives a similar 2.73-point deficit,
showing that both warmups contribute while joint answer learning supplies most
of the final task adaptation.

Figure~\ref{fig:result-panels}(c) shows that the curriculum changes are
concentrated on evidence-composition tasks rather than uniformly distributed.

\begin{table}[tp]
\centering
\caption{Architecture ablations on seen and unseen structures (accuracy, \%).}
\label{tab:arch-ablation}
\fontsize{9}{11}\selectfont
\renewcommand{\arraystretch}{1.18}
\setlength{\tabcolsep}{.5pt}
\begin{tabularx}{\linewidth}{@{}Z>{\centering\arraybackslash}p{23.5pt}>{\centering\arraybackslash}p{25.5pt}>{\centering\arraybackslash}p{36pt}>{\centering\arraybackslash}p{28pt}>{\centering\arraybackslash}p{25.5pt}>{\centering\arraybackslash}p{36pt}>{\centering\arraybackslash}p{24.5pt}!{\hspace{7pt}}>{\centering\arraybackslash}p{23.5pt}>{\centering\arraybackslash}p{25.5pt}>{\centering\arraybackslash}p{36pt}>{\centering\arraybackslash}p{28pt}>{\centering\arraybackslash}p{25.5pt}>{\centering\arraybackslash}p{36pt}>{\centering\arraybackslash}p{24.5pt}@{}}
\toprule
\rowcolor{wfheader}
 & \multicolumn{7}{c}{\cellcolor{wfopen}\textbf{(a) Seen structures}} & \multicolumn{7}{c}{\cellcolor{wfapi}\textbf{(b) Unseen structures}} \\
\cmidrule(lr){2-8}\cmidrule(lr){9-15}
\rowcolor{wfheader}
\textbf{Task} & \cellcolor{wfhuman}\wfhead{\textbf{QA}\\\textbf{only}} & \cellcolor{wfapi}\wfhead{\textbf{+}\\\textbf{Entity}} & \cellcolor{wfopen}\wfhead{\textbf{+}\\\textbf{Coverage}} & \cellcolor{wfhuman}\wfhead{\textbf{+}\\\textbf{Lattice}} & \cellcolor{wfred}\wfhead{\textbf{No}\\\textbf{Entity}} & \cellcolor{wfgreen}\wfhead{\textbf{No}\\\textbf{Coverage}} & \cellcolor{wfgreenhead}\wfhead{\textbf{Full}\\\textbf{model}} & \cellcolor{wfhuman}\wfhead{\textbf{QA}\\\textbf{only}} & \cellcolor{wfapi}\wfhead{\textbf{+}\\\textbf{Entity}} & \cellcolor{wfopen}\wfhead{\textbf{+}\\\textbf{Coverage}} & \cellcolor{wfhuman}\wfhead{\textbf{+}\\\textbf{Lattice}} & \cellcolor{wfred}\wfhead{\textbf{No}\\\textbf{Entity}} & \cellcolor{wfgreen}\wfhead{\textbf{No}\\\textbf{Coverage}} & \cellcolor{wfgreenhead}\wfhead{\textbf{Full}\\\textbf{model}} \\
\midrule
Claim status & 52.50 & 52.00 & 53.50 & 66.00 & 71.50 & 69.00 & \cellcolor{wfgreen}\best{83.00} & 34.00 & 33.67 & 33.33 & 43.33 & 39.67 & 37.33 & \cellcolor{wfgreen}\best{48.00} \\
Claim select. & 48.00 & 48.50 & 47.50 & 59.00 & 68.00 & 65.00 & \cellcolor{wfgreen}\best{80.50} & 30.00 & 30.33 & 30.67 & 36.67 & 34.67 & 31.33 & \cellcolor{wfgreen}\best{38.67} \\
Strongest & 58.00 & 57.00 & 58.50 & 77.00 & 87.00 & 84.50 & \cellcolor{wfgreen}\best{96.00} & 50.00 & 49.33 & 50.33 & 56.67 & 54.00 & 47.33 & \cellcolor{wfgreen}\best{58.67} \\
Resolve view & 28.50 & 29.00 & 27.50 & 48.00 & 61.00 & 59.50 & \cellcolor{wfgreen}\best{69.00} & 30.00 & 30.33 & 30.67 & 46.67 & 47.33 & 43.33 & \cellcolor{wfgreen}\best{55.33} \\
Necess./red. & 35.00 & 34.50 & 36.00 & 59.50 & 75.00 & 70.50 & \cellcolor{wfgreen}\best{84.00} & 23.33 & 23.67 & 23.00 & 30.67 & 26.00 & 22.33 & \cellcolor{wfgreen}\best{33.33} \\
Shortest seq. & 39.00 & 40.00 & 38.50 & 65.00 & 88.50 & 87.00 & \cellcolor{wfgreen}\best{97.50} & 40.00 & 39.67 & 40.67 & 58.33 & 58.67 & 57.33 & \cellcolor{wfgreen}\best{65.33} \\
Minimal set & 17.00 & 16.50 & 18.00 & 36.00 & 49.50 & 46.00 & \cellcolor{wfgreen}\best{59.50} & 10.00 & 10.33 & 9.67 & 23.33 & 27.33 & 24.67 & \cellcolor{wfgreen}\best{32.33} \\
Best count & 67.50 & 68.00 & 67.00 & 90.00 & 85.00 & 97.50 & \cellcolor{wfgreen}\best{98.00} & 55.00 & 54.67 & 55.33 & 60.67 & 55.67 & 60.67 & \cellcolor{wfgreen}\best{61.33} \\
First gain & 71.00 & 71.50 & 71.50 & 94.00 & 87.50 & 97.00 & \cellcolor{wfgreen}\best{97.50} & 61.67 & 62.00 & 61.00 & 73.33 & 62.67 & 76.00 & \cellcolor{wfgreen}\best{76.67} \\
Max. answer & 36.00 & 35.50 & 36.50 & 55.00 & 68.00 & 64.00 & \cellcolor{wfgreen}\best{87.00} & 22.00 & 21.67 & 22.33 & 30.00 & 28.00 & 24.00 & \cellcolor{wfgreen}\best{34.67} \\
\midrule
\textbf{Overall} & \cellcolor{wfred}45.25 & 45.25 & 45.45 & 64.95 & 74.10 & 74.00 & \cellcolor{wfgreen}\best{85.20} & \cellcolor{wfred}35.60 & 35.57 & 35.70 & 45.97 & 43.40 & 42.43 & \cellcolor{wfgreen}\best{50.43} \\
\bottomrule
\end{tabularx}
\par\smallskip
{\small\raggedright Task labels are shortened from Table~\ref{tab:worldscope-tasks}; Necess./red. denotes view necessity and redundancy. QA-only is the answer-training baseline. The three $+$ columns add entity fusion, coverage prediction, and the image-subset lattice cumulatively. Full model also uses counterworld supervision; No Entity and No Coverage remove the respective branch.\par}
\end{table}

\begin{table}[tp]
\centering
\caption{Curriculum ablations on seen and unseen structures (accuracy, \%).}
\label{tab:curriculum-ablation}
\fontsize{10}{12}\selectfont
\renewcommand{\arraystretch}{1.18}
\setlength{\tabcolsep}{1.4pt}
\begin{tabularx}{\linewidth}{@{}L{72pt}*{5}{Y}!{\hspace{7pt}}*{5}{Y}@{}}
\toprule
\rowcolor{wfheader}
 & \multicolumn{5}{c}{\cellcolor{wfopen}\textbf{(a) Seen structures}} & \multicolumn{5}{c}{\cellcolor{wfapi}\textbf{(b) Unseen structures}} \\
\cmidrule(lr){2-6}\cmidrule(lr){7-11}
\rowcolor{wfheader}
\textbf{Task} & \cellcolor{wfred}\wfhead{\textbf{No}\\\textbf{Stage 1}} & \cellcolor{wfred}\wfhead{\textbf{No}\\\textbf{warmups}} & \cellcolor{wfhuman}\wfhead{\textbf{QA-only}\\\textbf{Stage 3}} & \cellcolor{wfapi}\wfhead{\textbf{No}\\\textbf{Stage 2}} & \cellcolor{wfgreenhead}\wfhead{\textbf{Full}\\\textbf{model}} & \cellcolor{wfred}\wfhead{\textbf{No}\\\textbf{Stage 1}} & \cellcolor{wfred}\wfhead{\textbf{No}\\\textbf{warmups}} & \cellcolor{wfhuman}\wfhead{\textbf{QA-only}\\\textbf{Stage 3}} & \cellcolor{wfapi}\wfhead{\textbf{No}\\\textbf{Stage 2}} & \cellcolor{wfgreenhead}\wfhead{\textbf{Full}\\\textbf{model}} \\
\midrule
Claim status & 57.00 & 53.50 & \best{84.00} & 81.50 & \cellcolor{wfgreen}83.00 & 34.67 & 34.33 & 45.33 & 46.67 & \cellcolor{wfgreen}\best{48.00} \\
Claim select. & 50.00 & 49.00 & 75.50 & 78.00 & \cellcolor{wfgreen}\best{80.50} & 30.33 & 30.67 & 35.33 & 37.33 & \cellcolor{wfgreen}\best{38.67} \\
Strongest claim & 66.00 & 59.50 & 91.50 & 94.00 & \cellcolor{wfgreen}\best{96.00} & 53.33 & 49.67 & 54.00 & 56.00 & \cellcolor{wfgreen}\best{58.67} \\
Resolve view & 33.00 & 29.50 & 64.50 & 64.00 & \cellcolor{wfgreen}\best{69.00} & 31.67 & 30.67 & 51.00 & 50.00 & \cellcolor{wfgreen}\best{55.33} \\
Necess./red. & 37.50 & 36.00 & 79.50 & 79.00 & \cellcolor{wfgreen}\best{84.00} & 27.33 & 24.00 & 30.33 & 28.67 & \cellcolor{wfgreen}\best{33.33} \\
Shortest seq. & 45.00 & 40.00 & 93.50 & 92.00 & \cellcolor{wfgreen}\best{97.50} & 45.00 & 41.00 & 62.67 & 60.67 & \cellcolor{wfgreen}\best{65.33} \\
Minimal set & 25.00 & 17.50 & 56.50 & 55.00 & \cellcolor{wfgreen}\best{59.50} & 18.33 & 10.67 & 29.33 & 28.00 & \cellcolor{wfgreen}\best{32.33} \\
Best count & 82.50 & 70.00 & 95.50 & \best{98.50} & \cellcolor{wfgreen}98.00 & 56.67 & 55.67 & 60.67 & \best{61.67} & \cellcolor{wfgreen}61.33 \\
First gain & 82.50 & 72.50 & 97.00 & \best{98.00} & \cellcolor{wfgreen}97.50 & 68.33 & 63.33 & 75.33 & \best{77.00} & \cellcolor{wfgreen}76.67 \\
Max. answer & 45.00 & 36.00 & 82.00 & 84.00 & \cellcolor{wfgreen}\best{87.00} & 25.00 & 22.00 & 31.33 & 31.00 & \cellcolor{wfgreen}\best{34.67} \\
\midrule
\textbf{Overall} & 52.35 & \cellcolor{wfred}46.35 & 81.95 & 82.40 & \cellcolor{wfgreen}\best{85.20} & 39.07 & \cellcolor{wfred}36.20 & 47.53 & 47.70 & \cellcolor{wfgreen}\best{50.43} \\
\bottomrule
\end{tabularx}
\par\smallskip
{\small\raggedright Stage 1 grounds scene evidence; Stage 2 learns counterworld compatibility; Stage 3 jointly trains answers and evidence. No warmups omits both preliminary stages.\par}
\end{table}

The two warmup stages affect different tasks.  Without Stage~1, unseen resolving
view, shortest sequence, and minimal sufficient set fall by 23.66, 20.33, and
14.00 points, respectively, because the entity and coverage modules enter
compatibility learning without grounded scene evidence.  Removing Stage~2 has
smaller losses concentrated on resolving view, necessity/redundancy,
shortest sequence, and minimal sufficient set.  Its 0.33--0.34-point reversals on the
two count-view tasks correspond to one additional correct example and do not
form a consistent advantage.  Likewise, QA-only Stage~3 exceeds the full model
only on seen claim status; it remains lower on unseen claim status and on both
overall splits.

\subsection{Generalization and Remaining Difficulty}

The unseen-structure split remains substantially harder than the seen split:
full \worldflow{} obtains 50.43\% versus 85.20\%.  On unseen structures it
reaches 76.67\% on first count gain, 65.33\% on shortest sequence, and 61.33\%
on best count view.  Minimal sufficient set (32.33\%), necessity/redundancy
(33.33\%), and maximal identifiable answer (34.67\%) remain the hardest tasks.
These outputs require an exact set or several mutually consistent fields, so a
single missing component makes the complete prediction incorrect.  The
consistent drops after removing coverage, the evidence lattice, or Stage~2
show that evidence composition and judgments of sufficiency remain key sources
of error.

\begin{figure}[t]
  \centering
  \includegraphics[width=\textwidth]{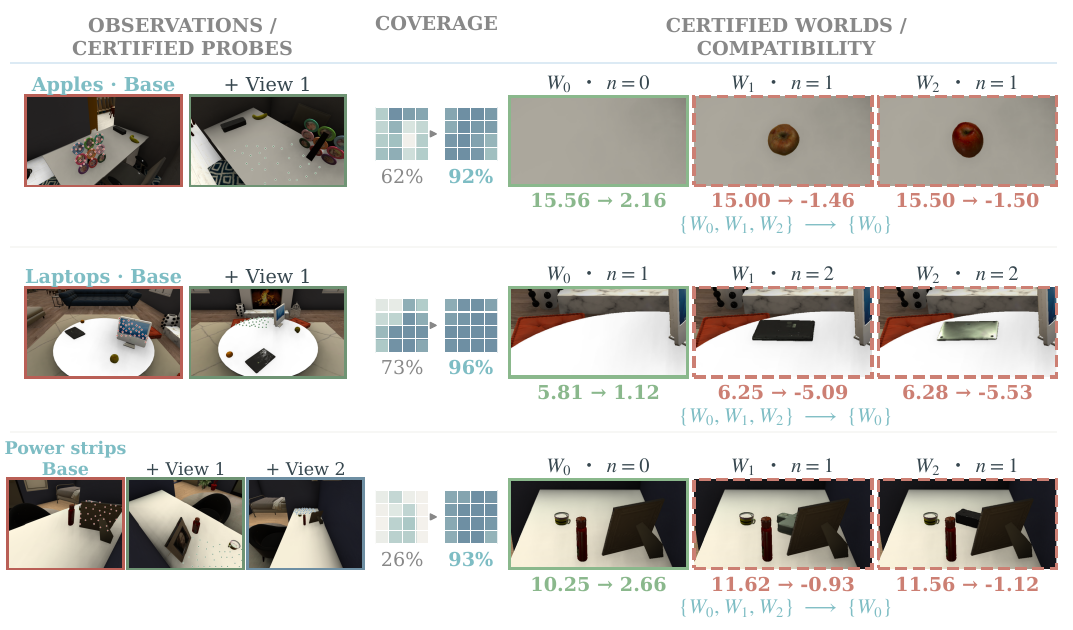}
  \caption{Internal evidence updates for three certified counterworld groups.
  Left: the base observation and one or two added views.  Colored dots show
  certified support-surface probes hidden from the base (red) and exposed by
  the added views (green and blue); these annotations are used only to
  visualize the offline supervision.  Middle: the model's $4\times4$
  support-surface coverage prediction before and after adding the views, with
  the mean value below each grid.  Right: the three certified complete worlds
  and the compatibility logit before $\rightarrow$ after the added evidence.
  Solid green retains the compatible world and dashed red marks worlds the
  model excludes.  $n$ is the true count of the queried category in each
  complete world.}
  \Description{Three rows of qualitative examples.  Each row shows a base and
  added RGB views with certified probes, a before-and-after four-by-four
  coverage grid, and three complete-world hypotheses whose compatibility
  scores contract after the added evidence.}
  \label{fig:worldflow-evidence-cases}
\end{figure}

\subsection{Qualitative Evidence Updates}
\label{sec:qualitative-evidence}

Figure~\ref{fig:worldflow-evidence-cases} traces the internal evidence update
for three certified groups from the selected full-model checkpoint.  The
colored points are physical support-surface probes from the offline
certificate: red points are hidden in the base observation, while green and
blue points mark portions revealed by added views.  They visualize the target
geometry and never enter the student input.  The adjacent $4\times4$ arrays
are the model's predicted coverage in canonical support-surface coordinates;
they are not a partition of image pixels.  Mean predicted coverage rises from
62\% to 92\% for apples, from 73\% to 96\% for laptops, and from 26\% to 93\%
for power strips as the certified blind region is exposed.

For this diagnostic only, we retain the training-time compatibility head and
query it against the certified complete-world representations.  Neither the
head nor these privileged representations contributes to answer generation.
Before the reveal, all three certified worlds have positive compatibility
logits in every row.  After the reveal, the logit of the observed completion
remains positive (2.16, 1.12, and 2.66), while both alternatives cross below
zero.  The three rows also include a two-view and a three-view observation set.
Their shared behavior shows that the subset state can accumulate complementary
views, expand physical-surface coverage, and contract the certified
alternatives in the direction required by the training objective.  These cases
are explanatory examples; the complete test results and ablations provide the
aggregate evidence.

\section{Conclusion}
\label{sec:conclusion}

\paragraph{Scope and limitations.}
\worldscope{} deliberately isolates partial-observation reasoning in
controlled tabletop scenes.  Its 38 target categories, support-surface
queries, rendered HSSD interiors, and finite camera families do not cover
deformable objects, articulated interactions, dynamic events, outdoor scenes,
or real sensor artifacts.  The structure-disjoint test measures transfer to
new layouts within this controlled domain; it does not establish transfer to
arbitrary photographs or embodied environments.

The certified counterworlds provide concrete witnesses of ambiguity, not an
exhaustive representation of all physically possible completions.  Each group
contains three complete worlds, which is sufficient to show that a base
observation can admit different answers and to supervise view-induced
exclusions.  Agreement among the three worlds is not treated as proof that no
other completion exists.  WorldFlow also compresses unresolved support space
into a category-neutral $4\times4$ map.  This representation is efficient, but
it can discard fine geometry and object-specific placement information that the
denser data executor uses to construct gold evidence.

Finally, the reported large-model variants are single training runs selected
on a fixed development set.  Exact-set scoring requires the predicted set to
match the complete gold set, so a prediction with one missing view receives no credit.
The substantial gap between seen and unseen
structures, especially on minimal sets and maximal identifiable answers,
indicates that generalization remains a major challenge.

\paragraph{Conclusion and future work.}
We introduced \worldscope{} to study world modeling under partial observation
by relating visual evidence to compatible possible worlds and the strongest
conclusions they jointly support.
\worldset{} supplies 1.2 million questions over shared evidence semantics,
certified counterworlds supervise how added observations exclude complete
worlds, and \worldbench{} evaluates ten ways of judging and composing partial
evidence.  \worldflow{} combines cross-view entity evidence and physical-surface
coverage into image-subset representations, using counterworld supervision to
learn how additional observations constrain compatible worlds.
Its 64.34\% exact accuracy exceeds same-backbone QA-only
training by 24.88 points, while the architecture and curriculum ablations
localize the largest gains to image-subset comparison and evidence-boundary
reasoning.

Future work can extend the same formulation beyond controlled tabletops.  One
direction is to certify compatible worlds from real multi-view captures using
reconstruction or additional sensors only during annotation.  A second is to
replace the fixed coverage grid with an adaptive, category-conditioned surface
representation that preserves fine feasible placements without enumerating
them at inference.  Richer possible-world supervision could cover relations,
object state, and temporal events, while an active system could select the next
camera view according to the predicted contraction of the answer set.  These
extensions would support world representations that update as evidence is
acquired, enabling progressively more precise conclusions about partially
observed scenes.

\bibliographystyle{unsrtnat}
\bibliography{worldscope_worldflow_references}
\end{document}